\documentclass{article}

\usepackage{PRIMEarxiv}

\usepackage[utf8]{inputenc} 
\usepackage[T1]{fontenc}    
\usepackage{hyperref}       
\usepackage{url}            
\usepackage{booktabs}       
\usepackage{amsfonts}       
\usepackage{nicefrac}       
\usepackage{microtype}      
\usepackage{lipsum}
\usepackage{fancyhdr}       
\usepackage{graphicx}       
\graphicspath{{media/}}     
\usepackage{bm}

\title{Tracking the industrial growth of modern China with high-resolution panchromatic imagery: A sequential convolutional approach
}

\author{
  Ethan Brewer \\
  New York University \\
  \texttt{ethan.brewer@nyu.edu} \\
   \And
  Zhonghui Lv \\
  William \& Mary \\
  \texttt{zlv@wm.edu} \\
   \And
  Dan Runfola \\
  William \& Mary \\
  \texttt{danr@wm.edu} \\
}

\begin{document}
\maketitle

\begin{abstract}
Due to insufficient or difficult to obtain data on development in inaccessible regions, remote sensing data is an important tool for interested stakeholders to collect information on economic growth. To date, no studies have utilized deep learning to estimate industrial growth at the level of individual sites. In this study, we harness high-resolution panchromatic imagery to estimate development over time at 419 industrial sites in the People's Republic of China using a multi-tier computer vision framework. We present two methods for approximating development: (1) structural area coverage estimated through a Mask R-CNN segmentation algorithm, and (2) imputing development directly with visible \& infrared radiance from the Visible Infrared Imaging Radiometer Suite (VIIRS). Labels generated from these methods are comparatively evaluated and tested. On a dataset of 2,078 50 cm resolution images spanning 19 years, the results indicate that two dimensions of industrial development can be estimated using high-resolution daytime imagery, including (a) the total square meters of industrial development (average error of 0.021 $\textrm{km}^2$), and (b) the radiance of lights (average error of 9.8 $\mathrm{\frac{nW}{cm^{2}sr}}$). Trend analysis of the techniques reveal estimates from a Mask R-CNN-labeled CNN-LSTM track ground truth measurements most closely. The Mask R-CNN estimates positive growth at every site from the oldest image to the most recent, with an average change of 4,084 $\textrm{m}^2$.
\end{abstract}


\section{Introduction}

Satellite imagery analysis using deep learning methods, specifically convolutional neural networks (CNNs), has grown in popularity since 2012, with uses extending into the estimation of population \cite{pop_africa}, wealth \cite{yeh_africa_wellbeing}, poverty \cite{jean_poverty}, conflict \cite{goodman_conflict}, migration \cite{migratory_flows}, education \cite{testscores}, and infrastructure \cite{brewerroads}, among other applications \cite{eurosat,seth_diss,pyshore,brewer_thesis}. These techniques have broadly illustrated that harnessing satellites to remotely track development over time in otherwise data sparse regions is a potentially effective strategy \cite{burke_sustain}.
\par
One currently untested application of deep learning with satellite imagery is the identification and monitoring of industrial sites (e.g., factories, power plants, ports). The development of industrial sites is of broad interest, as it can serve as a proxy for everything from economic development \cite{indus_econ} to the projection of soft power \cite{runfola_fdi}. Because of its interrelationship with national security or proprietary corporate interests, information on such large-scale development is often undocumented or difficult to obtain openly by interested parties. This article focuses on testing our capability to automatically detect and monitor industrial sites within China using high-resolution panchromatic satellite imagery. Largely unrecorded in structured open source text information, the size and extent of industrial sites in China can be observed through routine or targeted satellite collection. From select sources, many locations appear, on average, at least yearly in cloud-free high-resolution imagery from satellite-based sensors over the past 15 years; some locations of interest have temporal granularity of as high as one day.
\par
To-date, no work has explored the use of machine learning methods trained on satellite imagery to estimate, and monitor over time, the development of particular economic industries at the scale of individual sites. In this work, the primary research question we seek to answer is, ``How accurately can a convolutional deep learning system estimate development at industrial sites from high-resolution panchromatic imagery''. This paper focuses on 2,078 images across 419 unique, known sites in the People's Republic of China from 2002 to 2021. Each image is labeled using a Mask R-CNN (MR-CNN) to estimate structure footprint coverage, based on a transfer learning parameter tuning approach with a subset of 182 manually digitized images. The MR-CNN's predictions on each of the 2,078 images are then used as labels for a CNN-LSTM (long short-term memory) to estimate the total area covered by structures. The ultimate goal of this study is to explore the potential of a technique that accepts as input a single satellite image, and achieves the task of estimating a single, total building footprint metric for that image (i.e., we do not seek to estimate the spatial location of buildings, only their total coverage). This approach enables users to identify regions where rapid industrialization (or de-industrialization) may be occurring for more detailed, qualitative analyses.
\par
Our paper is organized as follows: In ``Related Work'', we review the relevant literature on measuring development and economic output, including building detection, using satellite imagery and deep neural networks (DNNs). In ``Data \& methodology'', we discuss the data and technical approach we use in the experiment. We introduce our results in the following section, and finally provide an analysis and conclusion in the final two sections.

\section{Related work}
In this section we provide a brief overview of the use of deep learning models with satellite imagery for urban mapping and economic metric estimation, methods for building detection within satellite images, and related deep learning approaches to evaluating changes across time from satellite imagery.
\par
There is a large body of work seeking to classify land cover from satellite imagery, including industrial areas, with linear, support vector, and classification trees being commonly applied tools \cite{landuse_svm,landuse_indus}. In recent years, techniques for analyzing social and economic outcomes and metrics have predominately centered on the application of DNNs. In 2020, Bo Yu at el. used publicly available, daytime, 30-meter resolution Landsat-7 images and a multi-task deep learning framework to estimate economic statistics at the county-level in mainland China \cite{landsat_china}. In 2020, Yeh et al. utilized CNNs to estimate survey-based estimates of asset wealth across approximately 20,000 African villages from publicly-available multispectral satellite imagery, with their models able to explain 70\% of the variation in ground-measured village wealth in countries where the model was not trained \cite{yeh_africa_wellbeing}; this built on earlier efforts by Jean et al. \cite{jean_poverty}. In 2021, as part of SpaceNet Challenge 7, Adam Van Etten et al. explored multi-temporal urban development (in this case, buildings) using mid-resolution (4.0 meter) satellite imagery mosaics. 24 images per location were used over the span of two years (one image per month) covering 101 unique geographies. The top results from the challenge demonstrate segmentation methods that are able to track building footprint construction and demolition over time with an F1 rating of 0.45 (IOU $\geq$ 0.25) \cite{spacenet_urbandev}.
\par
Outside of these applications, object detection algorithms have been applied for detection and extraction of features of interest in satellite imagery including roads, buildings, and vehicles \cite{vehicle_det}. Several issues - such as varying size, background, and orientation of target objects - make the automatic detection of features in satellite imagery a challenging problem \cite{BREWERpois,progress_objdet_sat}. Over the past few years, U-Nets \cite{unet}, and variants of region-based convolutional neural networks (R-CNNs) \cite{mrcnn}, have become popular DNN-based approaches for object detection and segmentation in the greater computer vision community, including within the computer vision remote sensing (CVRS) domain. Some examples of the use of U-Nets in CVRS include building detection from high-resolution multispectral imagery \cite{prathap_bldgdet}, and building footprint and road detection within OpenStreetMap (OSM) fused with Sentinel-1 and 2 imagery \cite{ayala_bldgroad}. 
\par
In the present work, we leverage a specific approach called Mask R-CNN. This method generates bounding boxes and segmentation masks for each instance of an object in an image. Mask R-CNN extends Faster R-CNN \cite{faster_rcnn} by adding a branch for predicting segmentation masks on each region of interest (RoI) in parallel with the existing branch for classification and bounding box regression. The mask branch is a small fully-connected network \cite{fcn} applied to each RoI, predicting a segmentation mask at the pixel scale \cite{mrcnn}. Germane implementation examples in remote sensing include building extraction from RGB-PanSharpened images from the DeepGlobe building extraction challenge \cite{zhao_mrcnn}, and large-scale building extraction in high-resolution RGB aerial orthophoto mosaics of urban areas in Chile \cite{stiller_mrcnn}.
\par
More broadly, several studies in recent years have sought to use DNNs to capture temporal changes in features contained in satellite imagery. For example, in 2021, a U-Net coupled with LSTM \cite{lstm} was harnessed for urban change detection on various mid to high-resolution imagery datasets \cite{unet_lstm_urbanchange}. Also in 2021, Zhu et al. utilized convolutional LSTM models trained on WorldView-2 and QuickBird imagery for land cover classification, with their results highlighting the effectiveness of DNNs and LSTMs for classifying complex land use and land cover maps with multitemporal high-resolution images \cite{conv_lstm_landuse}. Our work builds on these approaches, focusing on multi-temporal classification of industrial development using only a single band.

\section{Data \& methodology}
The methodology and data utilized in this study are summarized as follows:

\begin{enumerate}
    \item A Mask R-CNN is trained to measure the total area covered by structures using building location data from Shanghai ($N=4,582$) and fine-tuned with 182 high-resolution ($\approx$0.5 m; total of 545 million building pixels and 6.62 billion non-building) satellite images at known factories, power stations, and ports in China
    \item The MR-CNN results are used to label 2,078 high-resolution ($\approx$0.5 m) images
    \item 1,822 of the resulting image-label pairs are used train a CNN-LSTM, with the goal of accurately estimating the total footprint of development at an industrial site over time (but, notably, not the explicit pixel locations of that development)
    \item The effectiveness of the CNN-LSTM in estimating total structure area is tested on the remaining 256 images
    \item Results are compared to training the CNN-LSTM using low-resolution satellite-derived nighttime light (radiance) values as labels, serving as a secondary proxy for development
\end{enumerate}

\begin{figure}[!ht]
    \centering
    \includegraphics[width=350pt]{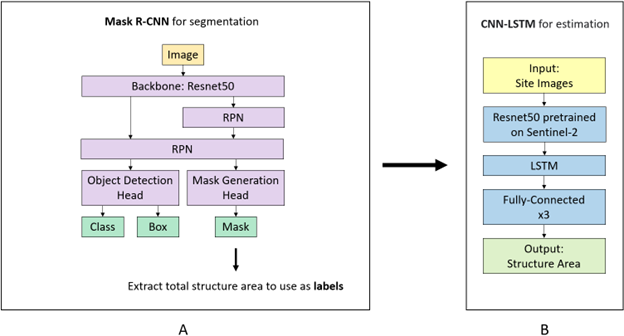}
    \caption{The overall method is to use a fine-tuned object detection algorithm to generate labels for a CNN-LSTM estimation model of structural footprint area.}
    \label{fig:overall_meth}
\end{figure}

\noindent
An overview of the deep learning methodology and architectures are illustrated in Figure \ref{fig:overall_meth}.

\begin{figure}[!ht]
    \centering
    \includegraphics[width=250pt]{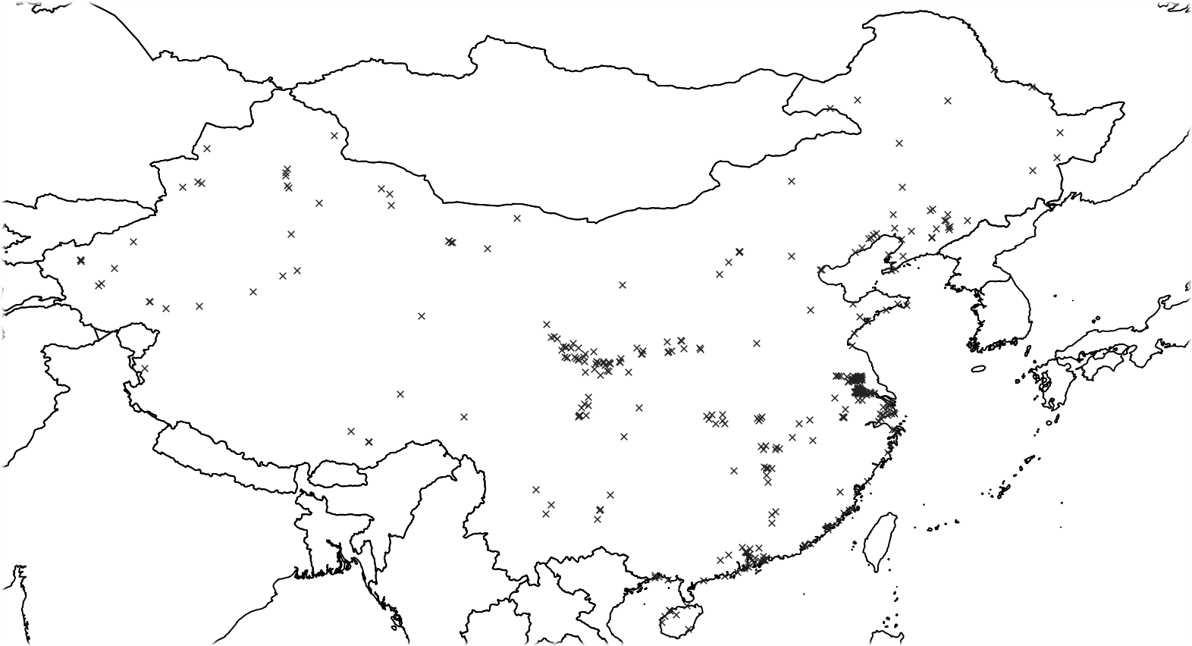}
    \caption{Factory, power station, and port locations, derived from GeoNames. Basemap is from \url{geoBoundaries.org} \cite{geoboundaries}.}
    \label{fig:sitemap}
\end{figure}

Sites of interest were selected using the GeoNames database \cite{geonames} as of September 2021, searching for ``factory'', ``power station'', and ``port'' within the country of China to filter results. This resulted in a sample of 419 industrial sites, encompassing 215 factories, 148 power stations (hydroelectric damns, coal plants, converter stations, etc.), and 56 ports. As seen in Figure \ref{fig:sitemap}, the sites cover a diverse geographic range throughout China's mainland borders. For each of these locations, image scenes were retrieved from DigitalGlobe \cite{DG}, owned and operated by Maxar Technologies. Image tiles were selected from DigitalGlobe based on the oldest archived strip available, spacing instances evenly to the newest strip, limiting cloud cover, and minimizing nadir. Images were then cropped from the tiles based on a 800x800 meter (64,000 $\textrm{m}^2$) geographic square centered around the location's coordinates.

\begin{figure}[!ht]
    \centering
    \includegraphics[width=350pt]{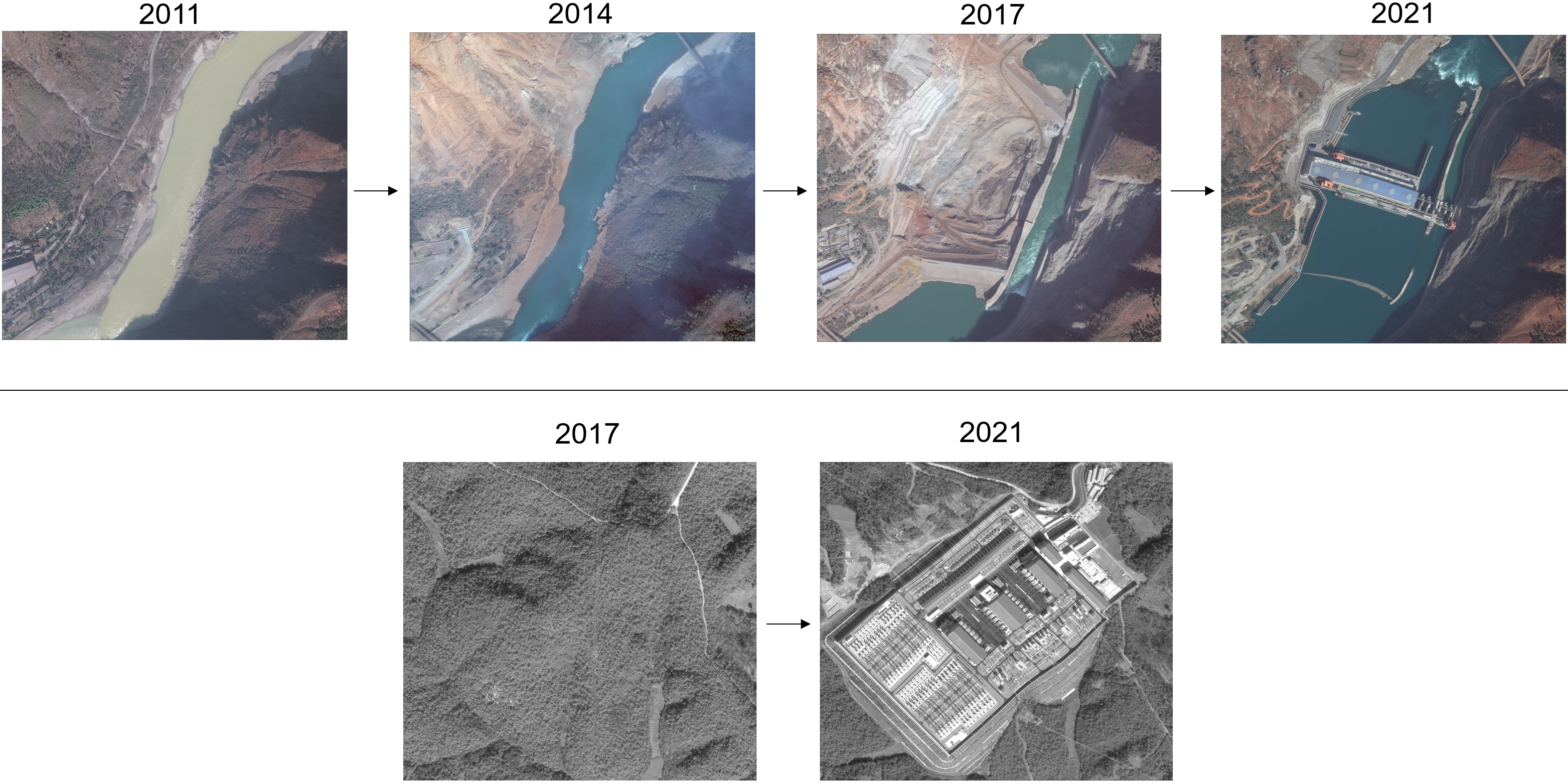}
    \caption{Two examples of rapid construction over a short period of time. The top row shows the development of the Jinsha hydroelectric power station in Sichuan province and one of the 103 RGB images in the 2,078 image dataset. The bottom row shows the Wudongde-Kunbei high-voltage, direct current (HVDC) static inverter plant in Yunnan province near Kunming. Images ©20[11,14,17,21], Maxar, USG Plus.}
    \label{fig:rapid_examples}
\end{figure}

\begin{figure}[!ht]
    \centering
    \includegraphics[width=200pt]{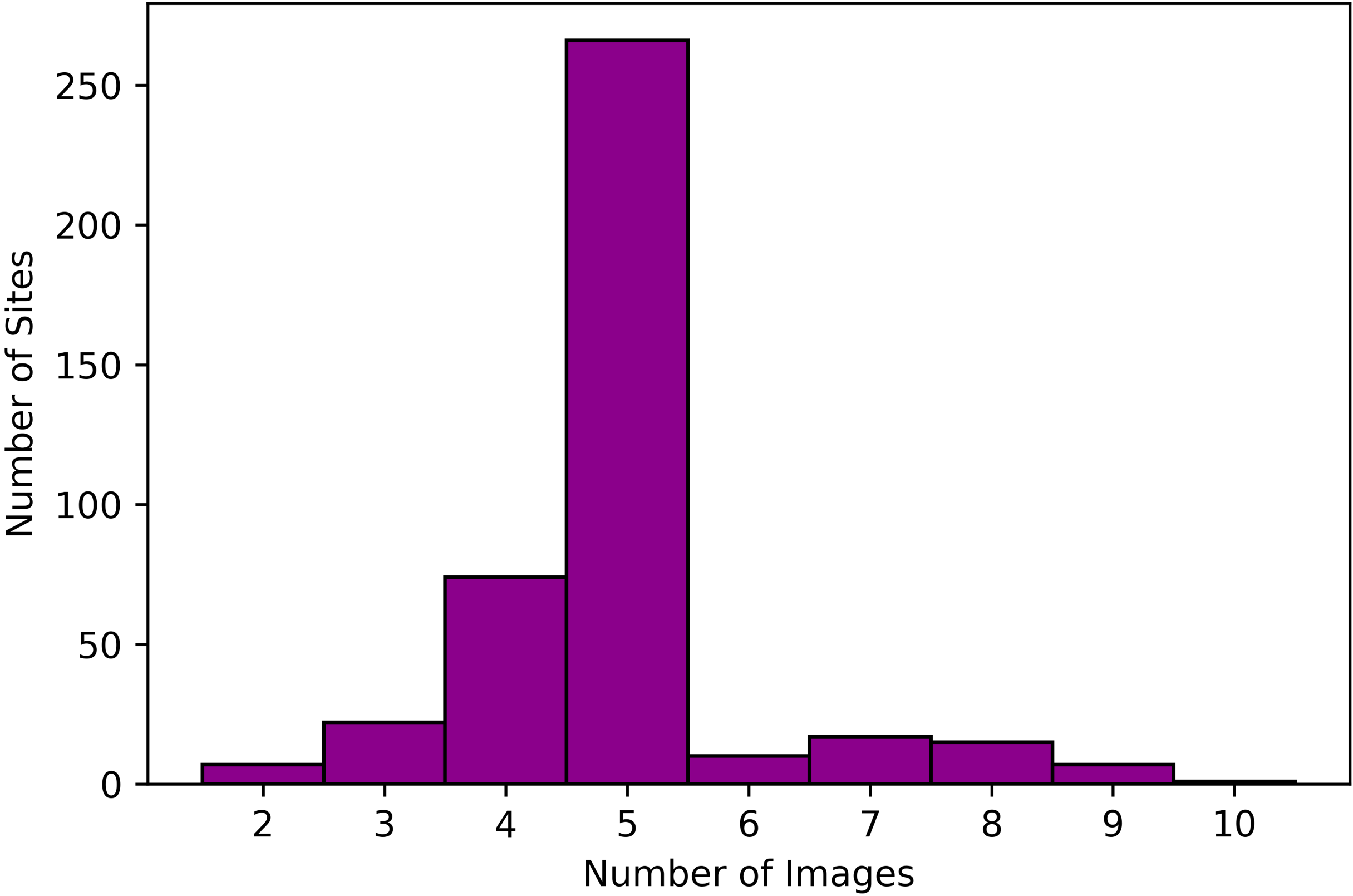}
    \caption{Histogram showing the number of sites (y-axis) that have the amount of images listed on the x-axis. 64\% of sites have 5 images through time.}
    \label{fig:sitenums}
\end{figure}

\begin{figure}[!ht]
    \centering
    \includegraphics[width=300pt]{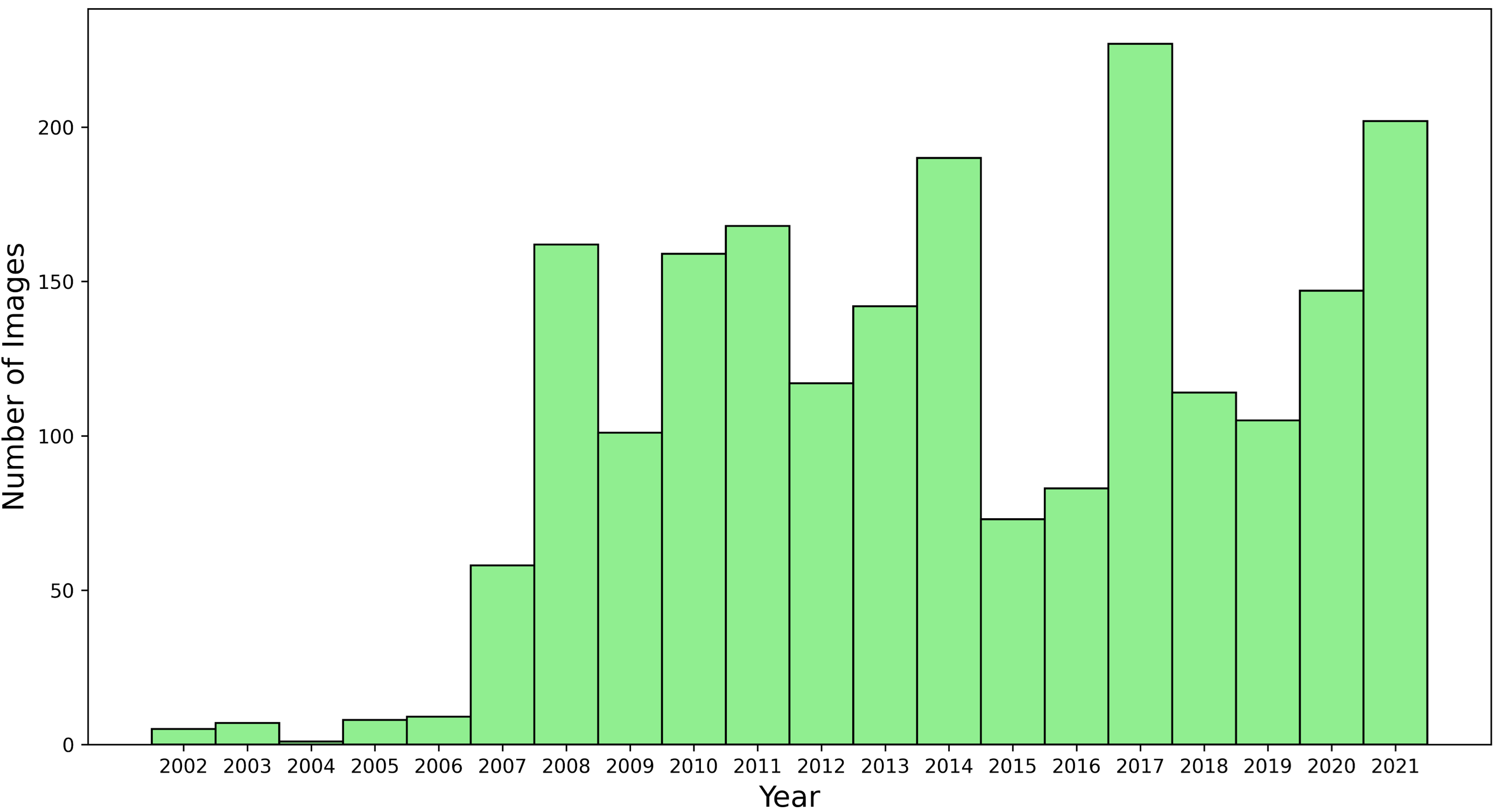}
    \caption{Histogram showing the number of images (y-axis) in each year listed on the x-axis. 99\% of the images are from 2007 to 2021.}
    \label{fig:sitedates}
\end{figure}

There are 2,078 total images covering 419 unique sites. Figure \ref{fig:rapid_examples} displays selections from two sites that underwent rapid growth. Figure \ref{fig:sitenums} shows the number of sites, with a median of 5 observations per site. These observations are relatively evenly distributed over time, with the number of instances per year shown in Figure \ref{fig:sitedates}. Image resolution varies from 30 to 60 cm, with a median resolution of 50 cm. 95\% of the images are single-band panchromatic, while the remaining 5\% are RGB. For input into the models, the RGB images were converted to single-band black and white images using the National Television System Committee (NTSC) luminance conversion formula \cite{luminance_convert}:

\begin{equation}
    Y = 0.299R+0.587G +0.114B
\end{equation}

\noindent
where for a given pixel with components normalized 0 to 1, $R$ is the red component, $B$ is the blue component, $G$ is the green component, and $Y$ is the final converted luminance (also 0 to 1).

\begin{figure}[!ht]
    \centering
    \includegraphics[width=425pt]{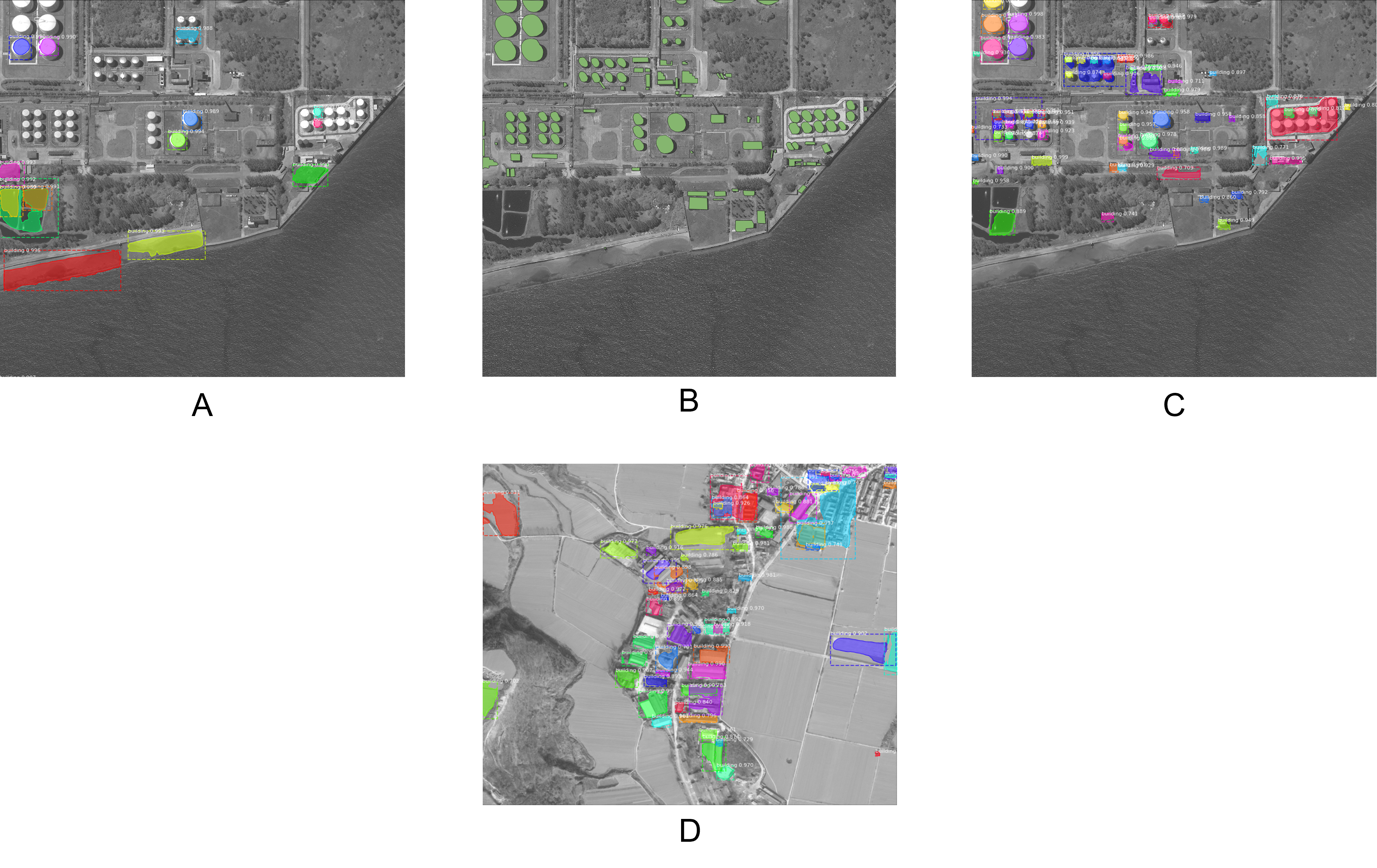}
    \caption{Visualization of (A) the performance of the Mask R-CNN after training on the Shanghai dataset before fine-tuning, (B) geocoding of structures at the site, and (C) the performance of the MR-CNN after fine-tuning on the geocoded set. (D) shows prediction on a random image in the DigitalGlobe dataset not included in the fine-tuning. For reference, the total estimated structural area for (C) is 32,447 $\textrm{m}^2$ and 54,973 $\textrm{m}^2$ for (D). Images ©20[17,21], Maxar, USG Plus.}
    \label{fig:mrcnn_performance}
\end{figure}

To train the Mask R-CNN, first the algorithm (pre-trained on MS COCO (Microsoft Common Objects in Context) dataset \cite{coco}) is trained on 4,582 30 cm resolution panchromatic Worldview-3 images of Shanghai, China and their associated building footprint masks \cite{spacenet_shanghai}. The images were upscaled from 640x640 pixels to 1024x1024. Each image depicts an area of 200x200 meters. Our MR-CNN hyperparameters were adopted from a MR-CNN trained on RGB images in the same Shanghai dataset \cite{Mstfakts} with only the image dimensions and region proposal network (RPN) anchor scales changed (to accommodate the larger image size). The Figure \ref{fig:mrcnn_performance}A shows a typical example of the model's performance on a site in the DigitalGlobe dataset. At this stage, the model struggles to identify industrial-specific structures. Therefore, to improve detection performance, the MR-CNN is fine-tuned on 182 DigitalGlobe images (resized to 1024x1024) and their associated masks. Of the 182, 37 are held-out for validation. The masks are created by manually geocoding the structures contained in the images. The 182 images were selected from across the latitudinal spectrum of the dataset with the most recent example of each site chosen to maximize the number of structures available to geocode. Due to the frequent difficulty in distinguishing industrial structures from commercial and residential, all structures present in the images are geocoded---i.e., buildings, silos, warehouses, piers, dams, electrical power relays, and antennas, and not including roads and bare pavement. A visual example of a geocoding is shown in Figure \ref{fig:mrcnn_performance}B. The average precision, $AP$, at defined intersection over union (IoU) thresholds is used to evaluate the performance. The $AP$ with an IoU threshold, $T$, is defined as

\begin{equation}
    AP_{T} = \frac{\textrm{True Positive}}{\textrm{True Positive} + \textrm{False Positive}}
\end{equation}

\noindent
where prediction instances with an $\textrm{IoU}<T$ are false positives, and predictions with $\textrm{IoU}\geq T$ are true positives. IoU is defined as the area of overlap between the prediction and the mask divided by the combined area of the prediction and mask.

The total area, $A$, in square meters covered by structures in an image is calculated with

\begin{equation}
    A = (800*800)\frac{P_S}{H*W}
    \label{eq:area}
\end{equation}

\noindent
where $P_S$ is the number of pixels in the image classified as structure, $H$ and $W$ are the pixel height and width of the image, and $800*800$ is the total geographic area (in $\textrm{m}^2$) depicted in an image. Overlaps in prediction masks were accounted for and removed in the calculation of $P_S$. With labels now generated, the CNN-LSTM model can be trained.
\par
As shown in Figure \ref{fig:overall_meth}B, the backbone of the CNN-LSTM is a ResNet50 pre-trained on 10-band, 10x10 meter resolution Sentinel-2 satellite imagery \cite{torchgeo}. During training and testing, each batch input into the network contains all (and only) the images of a single site. For input into the CNN, each image is downscaled, through bilinear interpolation, to 516 pixel height and 426 pixel width. These values are derived from the median image dimensions divided by 3.75 (a number converged upon through trial and error with GPU memory capacity). A second experiment is conducted in which only the the least and most recent instances of each site are used, allowing for image dimensions of 1004x841. As illustrated in Figure \ref{fig:overall_meth}B, the CNN portion outputs to an LSTM, after which dimensions are reduced through three fully-connected layers. The final output represents the estimates of the total structural area for each instance of the site. Adam optimization and L1 loss are utilized to optimize parameters across the full architecture. A 75-12.5-12.5\% split is carried out, with 1556 images used in training, 256 in validation, and 256 in testing.
\par
\begin{figure}[!ht]
    \centering
    \includegraphics[width=425pt]{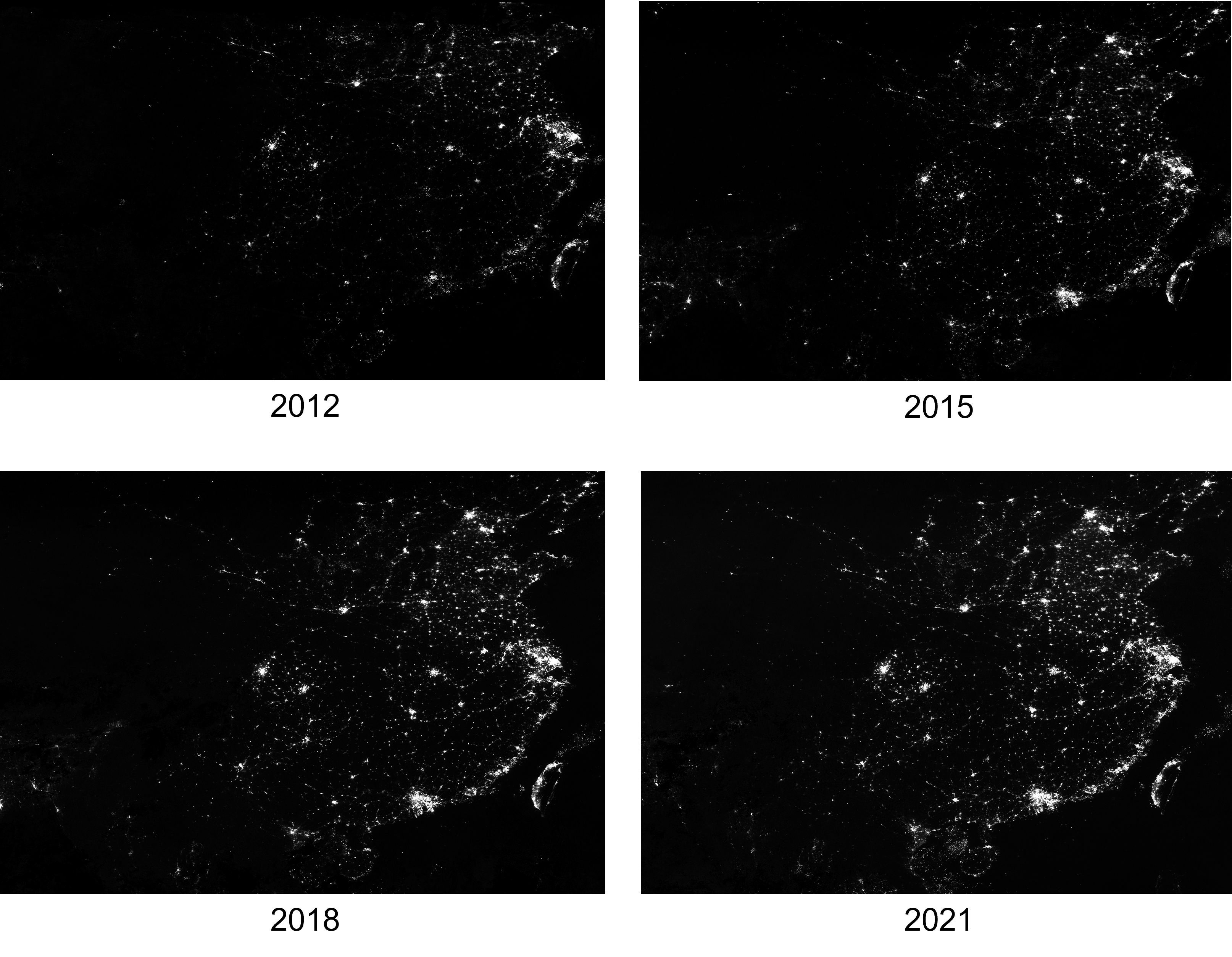}
    \caption{VIIRS composites covering China from 2012, 2015, 2018, and 2021. Notice the overall increase in radiance between 2012 and 2021 at this scale.}
    \label{fig:viirs}
\end{figure}

For a researcher interested in monitoring development of industrial sites, a simpler approach might be to examine nighttime light intensity. Nighttime lights (NTL) have proven to be accurate for urban mapping \cite{ntl_urbanmapping} and estimating wealth and poverty at city, state, and country levels \cite{jean_poverty, ni_ntl_poverty}. Here, we contrast the relative accuracy of using nighttime light radiance values with the CNN-LSTM to estimate development with the previously presented labeling approach. Our nighttime lights-derived strategy is based on the Visible Infrared Imaging Radiometer Suite (VIIRS) \cite{VIIRS}. Nightly, VIIRS collects visible and infrared global observations of Earth's land, atmosphere, and oceans with approximately 500x500 meter resolution at the equator (Figure \ref{fig:viirs}. The program was launched in April 2012, so only images in our dataset captured after this time will be analyzed by the model (1,367 total; 988 training; 183 validation; 196 testing). For our analysis, a nighttime light label is determined by finding the maximum pixel value (radiance) of the monthly average that overlaps with the daytime image crop. Only VIIRS pixels that have at least half of their area overlapping with the daytime image are selected.

To directly compare nighttime light estimations with those of the Mask R-CNN-based estimations, a linear model of the form

\begin{equation}
    A = m*NTL + b_0
    \label{eq:area_from_ntl}
\end{equation}
 
\noindent
is conceived to approximate structured area, $A$. $NTL$ are the predicted NTL values from the CNN-LSTM or raw NTL values, $m$ is the slope, and $b_0$ is the y-intercept. To build the linear model, values for $A$ are obtained from 181 ground truth, geocoded images (with one site excluded due to an image being older than 2012) through equation \ref{eq:area}, where $P_S$ is the number of hand-digitized pixels.

\section{Results}
Figure \ref{fig:mrcnn_performance}C shows the Mask R-CNN result on a geocoded image after fine-tuning. Figure \ref{fig:mrcnn_performance}D is a random result on an image not geocoded or included in fine-tuning. The average precision, $AP$, at various IoU thresholds on the 37 validation images is shown in Table \ref{tab:APs_T}. Table \ref{tab:APs_dates} shows the $AP$ at an IoU threshold of 0.3 grouped by image year on the entire 182-image geocoded dataset.

\begin{table*}[!ht]
    \begin{center}
    \begin{tabular}{|c|c|}
    \hline
      ${\bm{T}}$  & ${\bm{AP}}$ \\ 
      \hline
      0.1  & 0.30 \\ 
      \hline
      0.3 & 0.20 \\ 
      \hline
      0.5 & 0.11 \\ 
      \hline
    \end{tabular}
    \end{center}
    \caption{Average precision, $AP$, of the MR-CNN at IoU thresholds, $T$, of 0.1, 0.3, and 0.5.}
    \label{tab:APs_T}
\end{table*}

\begin{table*}[!ht]
    \begin{center}
    \begin{tabular}{|c|c|}
    \hline
    {\bf{Year}}  & ${\bm{AP}}$ \\
    \hline
    2018  & 0.20 \\ 
    \hline
    2019 & 0.19 \\ 
    \hline
    2020 & 0.17 \\ 
    \hline
    2021 & 0.22 \\ 
    \hline
    \end{tabular}
    \end{center}
    \caption{Average precision, $AP$, of the MR-CNN at IoU thresholds of 0.3 for geocoded images from 2018-2021. Years before 2018 were excluded due to there being less than 9 examples of each.}
    \label{tab:APs_dates}
\end{table*}

Using the process outlined in ``Data \& methodology'', the total area of structures contained in each image in the DigitalGlobe dataset is computed using the MR-CNN. Across the 2,078 images, the average structural area is 48,106 $\textrm{m}^2$ with a standard deviation of 27,753 $\textrm{m}^2$. The average difference between the most and least recent instance of each site is 4,894 $\textrm{m}^2$. This compares to an average structure area of 82,909 $\textrm{m}^2$ in the geocoded dataset, with a standard deviation of 72,963 $\textrm{m}^2$.
\par
\begin{figure}[!ht]
    \centering
    \includegraphics[width=250pt]{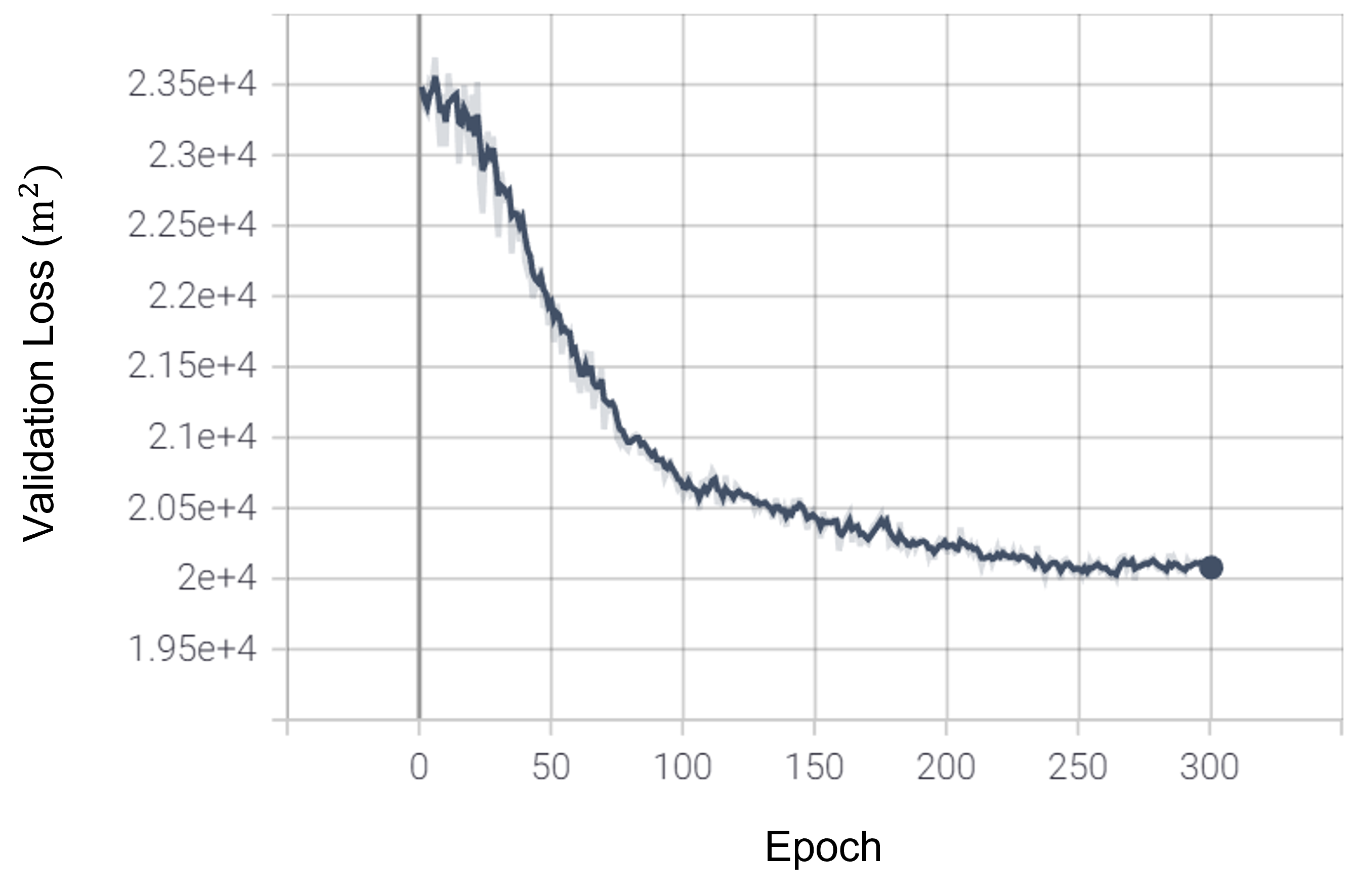}
    \caption{Validation loss vs. epoch during CNN-LSTM training on the entire dataset. The lowest loss occurs at epoch 271.}
    \label{fig:cnnlstm_val_loss}
\end{figure}

Training the CNN-LSTM on the entire DigitalGlobe dataset yields a test loss of 21,399 $\textrm{m}^2$. Figure \ref{fig:cnnlstm_val_loss} shows the validation loss trend during training. Training on only the least and most recent sites with larger images generates a slightly improved loss of 20,890 $\textrm{m}^2$.

Determination of nighttime light values, as outlined in ``Data \& methodology'', generates an average label of 13.0 $\mathrm{\frac{nW}{cm^{2}sr}}$ with a standard deviation of 21.5 $\mathrm{\frac{nW}{cm^{2}sr}}$. The average difference between the labels on the most and least recent instance of each site is 3.1 $\mathrm{\frac{nW}{cm^{2}sr}}$. Using these labels for CNN-LSTM training on the dataset produces an L1 loss of 9.8 $\mathrm{\frac{nW}{cm^{2}sr}}$ on the test dataset.
\par
\begin{figure}[!ht]
    \centering
    \includegraphics[width=425pt]{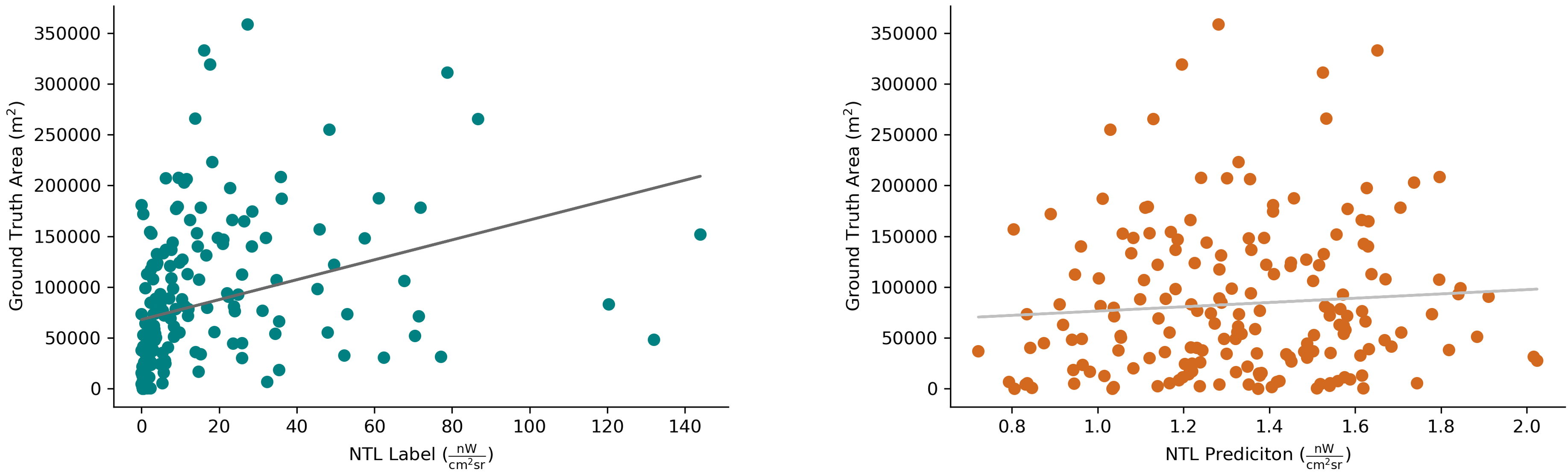}
    \caption{Results of linear regression performed (A) directly on the raw NTL labels, and (B) on the NTL predictions from the CNN-LSTM.}
    \label{fig:ntl_to_area}
\end{figure}

Figure \ref{fig:ntl_to_area} shows the the performance of NTL (raw labels and CNN-LSTM predictions) to approximate structural area through linear regression (equation \ref{eq:area_from_ntl}). The $R^2$ of the fit lines are 0.10 and 0.01 for Figs. \ref{fig:ntl_to_area}A and B, respectively.

\section{Analysis \& discussion}
\begin{figure}[!ht]
    \centering
    \includegraphics[width=300pt]{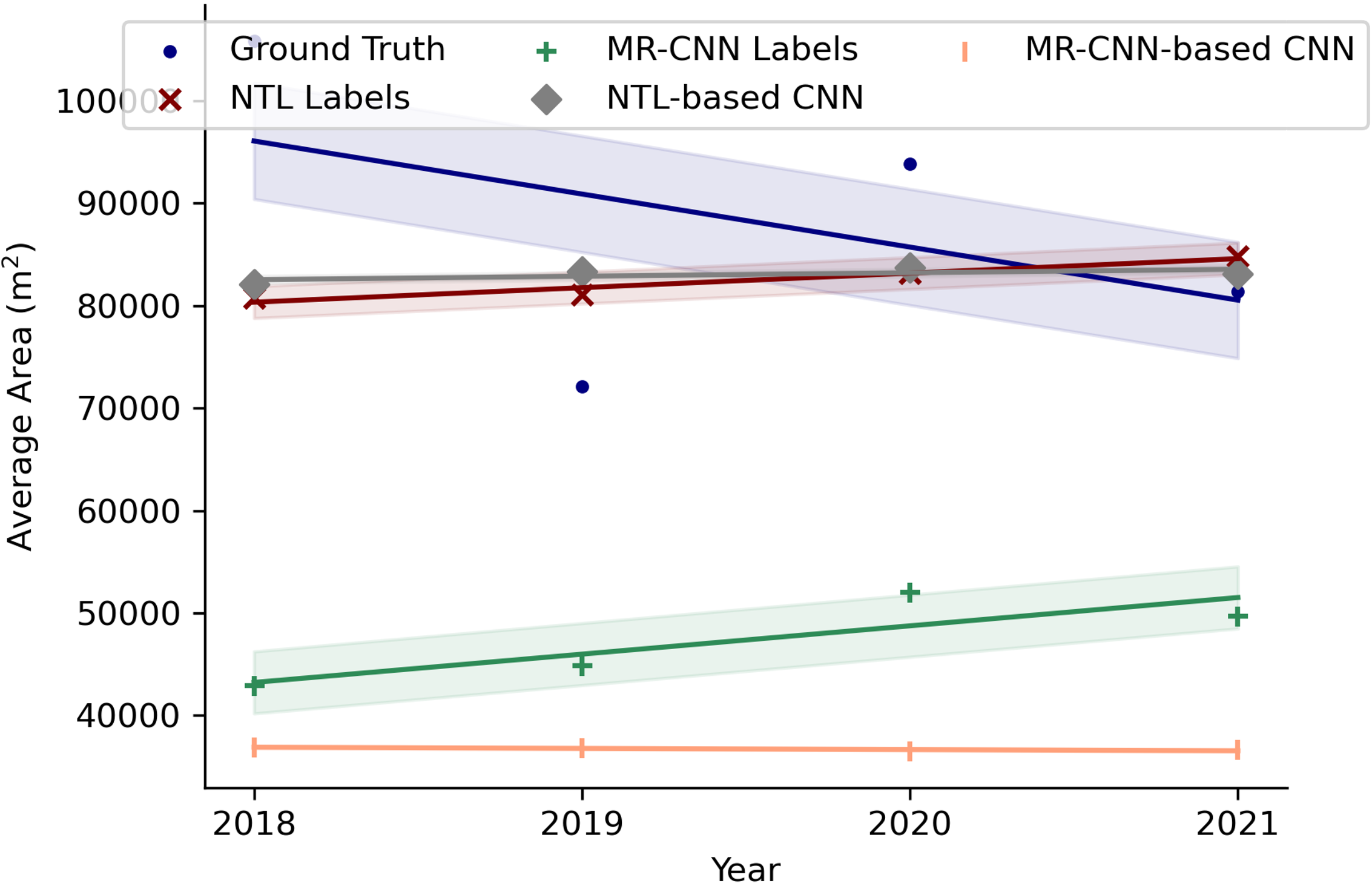}
    \caption{Average area estimations at the 181 geocoded sites from 2018 to 2021. The are 9 sites in 2018, 21 in 2019, 45 in 2020, and 94 in 2021. The band around each fit represents its 95\% confidence interval.}
    \label{fig:area_vs_year}
\end{figure}

To compare four area estimation strategies---(1) directly using raw NTL labels, (2) directly using raw MR-CNN labels, (3) MR-CNN-based CNN-LSTM, and (4) NTL-based CNN-LTSM---to ground truth, each technique is evaluated on the geocoded dataset from 2018 to 2021 (169 images corresponding to 169 sites). Figure \ref{fig:area_vs_year} displays the averaged estimations, grouped by year. Table \ref{tab:area_slopes} shows the percent change in area estimated by each technique. 
\par
Of particular note for our application is that - across our ground truth dataset - the total amount of land that was developed decreased. This is likely reflective of highly publicized industrial greening efforts being undertaken nationwide \cite{shanghai_tearline}. As we are most interested in detecting these types of trends, it is important that the algorithm used for change detection is capable of identifying the correct trend (i.e., we focus on our ability to accurately predict the trend of development, not necessarily the accuracy of any individual absolute square meterage of developed land measurement). The average trends predicted are shown in figure 4.12, by technique; of particular note is that the Mask R-CNN-based CNN-LSTM method is the only method that predicts the decline contained in the ground truth.

\begin{table*}[!ht]
    \begin{center}
    \begin{tabular}{|l|c|c|}
    \hline
    {\bf{Technique}}  & \bf{\% change per year} & \bf{\% change from 2018-2021} \\
    \hline
    Ground Truth & -7.5\% & -22.5\% \\ 
    \hline
    NTL Labels & +2.0\% & +6.0\% \\ 
    \hline
    MR-CNN Labels & +4.0\% & +12.0\% \\ 
    \hline
    NTL-based CNN & +0.5\% & +1.5\% \\
    \hline
    MR-CNN-based CNN & -0.2\% & -0.6\% \\
    \hline
    \end{tabular}
    \end{center}
    \caption{The \% area change of each estimation technique on the 169 geocoded images covering 2018 to 2021.}
    \label{tab:area_slopes}
\end{table*}

\begin{figure}[!ht]
    \centering
    \includegraphics[width=425pt]{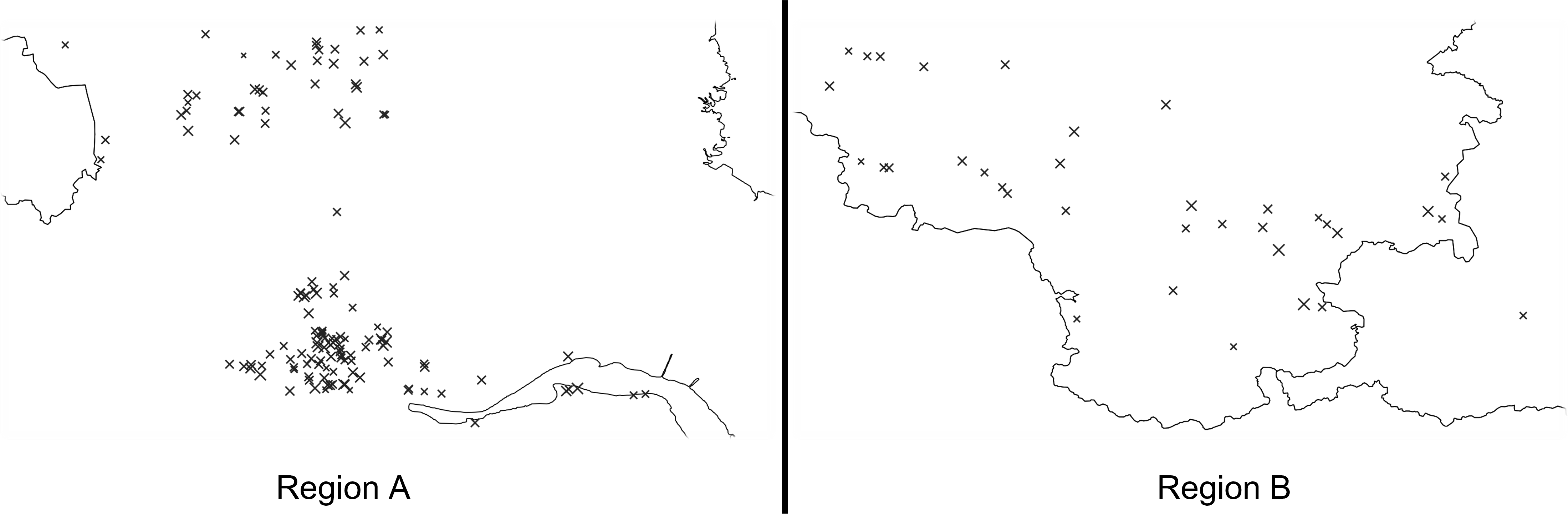}
    \caption{Region (A) shows an industrial region in Jiangsu province (near Shanghai). Region (B) shows an area in Gansu province. The marker sizes represent the relative change in the structural area estimated by the MR-CNN CNN-LSTM at each site from its oldest instance to its most recent. Basemap is from \url{geoBoundaries.org} \cite{geoboundaries}.}
    \label{fig:site_change}
\end{figure}

Using the Mask R-CNN-based CNN-LSTM on all 419 sites, Figure \ref{fig:site_change} shows the estimated relative development change in two regions in China from the oldest to the most recent instance of each site. Every site is estimated by the model as having growth in development. The average change is +4,084 $\textrm{m}^2$.

\subsection{Limitations}
While the Mask R-CNN CNN-LSTM was the only technique able to accurately predict the negative trend in developed industrial land, it also opens the door to numerous future inquiries. The Mask R-CNN precision results suggest limitations in the use of single-band satellite imagery with current state-of-the art segmentation models. Our precision values are in contrast to other examples of object detection from high-resolution satellite imagery where average precision values upwards of 0.94 at an IoU threshold of 0.5 have been reported \cite{maskrcnn_high_AP}. The relative ineffectiveness of the MR-CNN in this study most likely derives from the fact that the images contain a single band as opposed to the mutlispectral imagery most often examined in the literature. The differentiated color information encoded into multispectral imagery provides more information for the model to distinguish between similarly shaped features. For example, the inclusion of color is likely a key input in the estimation of buildings (grey, e.g.), ponds/lakes (blue), and fields/woodland (green). A common inaccuracy that occurs during inference is the prediction of fields, forests, and ponds as industrial structures.
\par
The labels and CNN-LSTM results derived from nighttime lights data show that 500x500 meter resolution VIIRS nighttime lights has difficulty at capturing differentiation in development at the sub-km scale. Highlighting this is the fact that every NTL prediction by the CNN-LSTM on the 181 image geocoded set fell under 2.02 $\mathrm{\frac{nW}{cm^{2}sr}}$ (close to the median), despite there being significantly more range to the underlying values, as seen in Figs. \ref{fig:ntl_to_area}A\&B. This suggests there are not significant features present in the images correlating with their respective nighttime light values.
\par
Finally, this study would benefit from greatly increasing the number of geocoded images, for the MR-CNN, and for technique trend analysis. Increasing the number of geocoded sites, such as geocoding two instances of each site, would provide more robust information for the MR-CNN to train on and, at least as importantly, allow for tracking the ground truth development at specific sites (to then contrast estimation techniques with).

\section{Conclusion}
In answering the research question ``How accurately can a convolutional deep learning system estimate development at industrial sites from high-resolution panchromatic imagery?'', the work presented in this paper provides three core contributions to the literature. First, we offer a comparison of the effectiveness of various techniques for remotely estimating development at industrial sites using deep learning and satellite data. Second, we discover that the resolution of current nighttime light sensors is generally insufficient to resolve development at the scale of individual industrial sites. Third, we provide evidence panchromatic imagery is relatively not well-suited for CVRS object detection tasks.
\par
A CNN-LSTM is able to resolve structural area and radiance to approximately 0.021 $\textrm{km}^2$ and 10 $\mathrm{\frac{nW}{cm^{2}sr}}$, respectively, at the tested industrial sites in China. The estimations from the NTL-based model approximate structural area with an $R^2$ of 0.01, with the raw labels alone approximating with an higher $R^2$ of 0.1 (Figure \ref{fig:ntl_to_area}). As seen in Figure \ref{fig:area_vs_year}, the hand-geocoded information reveals more recent images of industrial sites have less structural area on average, a result driven by small sample size or actual deindustrialization efforts by China. Mask R-CNN-based CNN-LSTM estimations reflect this trend, but NTL labels, MR-CNN labels, and NTL-based CNN-LSTM estimations do not. The labels generated by the Mask R-CNN, and the resulting predictions by the CNN-LSTM trained on those labels, indicate widespread growth over time when each site is tracked individually (average change of 4,084 $\textrm{m}^2$), as illustrated in Figure \ref{fig:site_change}.
\par
The contributions provided in this study provide meaningful future directions for related work. One such direction would be to test the methods detailed here with multispectral imagery, if available for these locations, or at locations where multispectral imagery is more readily accessible. Second, using the methods and imagery in this study, there may be significant improvements in MR-CNN (or related algorithm, e.g., U-Net) performance with a substantial addition to the number of carefully geocoded sites. A greater number of geocoded sites would also increase the sample size of ground truth observations allowing more robust trend analysis of the various prediction techniques.

\section*{Acknowledgments}
The authors would like to acknowledge Chris Rasmussen of the National Geospatial Agency (NGA) for their contribution in providing access to Maxar's DigitalGlobe archive. The authors also acknowledge the contributions of Michael Getaneh, John Hennin, Sidonie Horn, Tara McLaughlin, and Olivia Wachob of William \& Mary for their help in ordering the DigitalGlobe image strips.

Finally, the authors acknowledge William \& Mary Research Computing for providing computational resources and technical support that have contributed to the results reported within this paper. URL: \url{https://www.wm.edu/it/rc}.

\bibliographystyle{unsrt}  
\bibliography{references}

\begin{thebibliography}{10}

\bibitem{pop_africa}
Wenjie Hu, Jay~Harshadbhai Patel, Zoe-Alanah Robert, Paul Novosad, Samuel
  Asher, Zhongyi Tang, Marshall Burke, David Lobell, and Stefano Ermon.
\newblock Mapping missing population in rural india: A deep learning approach
  with satellite imagery.
\newblock In {\em Proceedings of the 2019 AAAI/ACM Conference on AI, Ethics,
  and Society}, AIES '19, page 353–359, New York, NY, USA, 2019. Association
  for Computing Machinery.

\bibitem{yeh_africa_wellbeing}
Christopher Yeh, Anthony Perez, Anne Driscoll, George Azzari, Zhongyi Tang,
  David Lobell, Stefano Ermon, and Marshall Burke.
\newblock Using publicly available satellite imagery and deep learning to
  understand economic well-being in africa.
\newblock {\em Nature Communications}, 11, 05 2020.

\bibitem{jean_poverty}
Neal Jean, Marshall Burke, Michael Xie, W.~Matthew Davis, David~B. Lobell, and
  Stefano Ermon.
\newblock Combining satellite imagery and machine learning to predict poverty.
\newblock {\em Science}, 353(6301):790--794, 2016.

\bibitem{goodman_conflict}
Seth Goodman, Ariel BenYishay, and Daniel Runfola.
\newblock A convolutional neural network approach to predict non-permissive
  environments from moderate-resolution imagery.
\newblock {\em Transactions in GIS}, 25(2):674--691, 2021.

\bibitem{migratory_flows}
Daniel Runfola, Heather Baier, Laura Mills, Maeve Naughton-Rockwell, and
  Anthony Stefanidis.
\newblock Deep learning fusion of satellite and social information to estimate
  human migratory flows.
\newblock {\em Transactions in GIS}, 2022.

\bibitem{testscores}
D.~Runfola, A.~Stefanidis, and H.~Baier.
\newblock Using satellite data and deep learning to estimate educational
  outcomes in data-sparse environments.
\newblock {\em Remote Sensing Letters}, 13(1):87--97, 2022.

\bibitem{brewerroads}
Ethan Brewer, Jason Lin, Peter Kemper, John Hennin, and Dan Runfola.
\newblock Predicting road quality using high resolution satellite imagery: A
  transfer learning approach.
\newblock {\em PLOS ONE}, 16(7):1--18, 07 2021.

\bibitem{eurosat}
Patrick Helber, Benjamin Bischke, Andreas Dengel, and Damian Borth.
\newblock Eurosat: A novel dataset and deep learning benchmark for land use and
  land cover classification.
\newblock {\em IEEE Journal of Selected Topics in Applied Earth Observations
  and Remote Sensing}, 12(7):2217--2226, 2019.

\bibitem{seth_diss}
Seth~M. Goodman.
\newblock {\em Filling in the Gaps: Applications of Deep Learning, Satellite
  Imagery, and High Performance Computing for the Estimation and Distribution
  of Geospatial Data}.
\newblock PhD thesis, William \& Mary, 2021.
\newblock Copyright - Database copyright ProQuest LLC; ProQuest does not claim
  copyright in the individual underlying works; Last updated - 2021-05-11.

\bibitem{pyshore}
Zhonghui Lv, Karinna Nunez, Ethan Brewer, and Dan Runfola.
\newblock pyshore: A deep learning toolkit for shoreline structure mapping with
  high-resolution orthographic imagery and convolutional neural networks.
\newblock {\em Computers \& Geosciences}, 171:105296, 2023.

\bibitem{brewer_thesis}
Ethan Brewer.
\newblock Deep learning from space: Methods \& applications in high-resolution
  satellite imagery analysis.
\newblock {\em William \& Mary Dissertations and Theses}, 2022.

\bibitem{burke_sustain}
Marshall Burke, Anne Driscoll, David~B. Lobell, and Stefano Ermon.
\newblock Using satellite imagery to understand and promote sustainable
  development.
\newblock {\em Science}, 371(6535):eabe8628, 2021.

\bibitem{indus_econ}
Canfei He, Zhiji Huang, and Rui Wang.
\newblock Land use change and economic growth in urban china: A structural
  equation analysis.
\newblock {\em Urban Studies}, 51(13):2880--2898, 2014.

\bibitem{runfola_fdi}
Jonas~B. Bunte, Harsh Desai, Kanio Gbala, Bradley Parks, and Daniel~Miller
  Runfola.
\newblock Natural resource sector fdi, government policy, and economic growth:
  Quasi-experimental evidence from liberia.
\newblock {\em World Development}, 107:151--162, 2018.

\bibitem{landuse_svm}
Arun Mondal, Sananda Kundu, Drsurendra Chandniha, Rituraj Shukla, and Prabhash
  Mishra.
\newblock Comparison of support vector machine and maximum likelihood
  classification technique using satellite imagery.
\newblock {\em International Journal of Remote Sensing and GIS}, 1:116--123, 01
  2012.

\bibitem{landuse_indus}
John Prince, Poovalingam Sivasubramanian, N.~Chandrasekar, and Durairaj K.S.P.
\newblock An analysis of land use pattern in the industrial development city
  using high resolution satellite imagery.
\newblock {\em Journal of Geographical Sciences}, 21:79--88, 02 2011.

\bibitem{landsat_china}
Bo~Yu, Ying Dong, Fang Chen, and Yu~Wang.
\newblock Economy estimation of mainland china at county-level based on landsat
  images and multi-task deep learning framework.
\newblock {\em Photogrammetric Engineering \& Remote Sensing}, 86(2):99--105,
  2020.

\bibitem{spacenet_urbandev}
Adam~Van Etten, Daniel Hogan, Jesus Martinez{-}Manso, Jacob Shermeyer, Nicholas
  Weir, and Ryan Lewis.
\newblock The multi-temporal urban development spacenet dataset.
\newblock {\em CoRR}, abs/2102.04420, 2021.

\bibitem{vehicle_det}
Xueyun Chen, Shiming Xiang, Cheng-Lin Liu, and Chun-Hong Pan.
\newblock Vehicle detection in satellite images by hybrid deep convolutional
  neural networks.
\newblock {\em IEEE Geoscience and Remote Sensing Letters}, 11(10):1797--1801,
  2014.

\bibitem{BREWERpois}
Ethan Brewer, Jason Lin, and Dan Runfola.
\newblock Susceptibility \& defense of satellite image-trained convolutional
  networks to backdoor attacks.
\newblock {\em Information Sciences}, 603:244--261, 2022.

\bibitem{progress_objdet_sat}
Kanchan Bhil, Rithvik Shindihatti, Shifa Mirza, Siddhi Latkar, Y.~S. Ingle,
  N.~F. Shaikh, I.~Prabu, and Satish~N. Pardeshi.
\newblock Recent progress in object detection in satellite imagery: A review.
\newblock In Sagaya Aurelia, Somashekhar~S. Hiremath, Karthikeyan Subramanian,
  and Saroj~Kr. Biswas, editors, {\em Sustainable Advanced Computing}, pages
  209--218, Singapore, 2022. Springer Singapore.

\bibitem{unet}
Olaf Ronneberger, Philipp Fischer, and Thomas Brox.
\newblock U-net: Convolutional networks for biomedical image segmentation.
\newblock {\em CoRR}, abs/1505.04597, 2015.

\bibitem{mrcnn}
Kaiming He, Georgia Gkioxari, Piotr Doll{\'{a}}r, and Ross~B. Girshick.
\newblock Mask {R-CNN}.
\newblock {\em CoRR}, abs/1703.06870, 2017.

\bibitem{prathap_bldgdet}
Geesara Prathap and Ilya Afanasyev.
\newblock Deep learning approach for building detection in satellite
  multispectral imagery.
\newblock In {\em 2018 International Conference on Intelligent Systems (IS)},
  pages 461--465, 2018.

\bibitem{ayala_bldgroad}
Christian Ayala, Rubén Sesma, Carlos Aranda, and Mikel Galar.
\newblock A deep learning approach to an enhanced building footprint and road
  detection in high-resolution satellite imagery.
\newblock {\em Remote Sensing}, 13(16), 2021.

\bibitem{faster_rcnn}
Shaoqing Ren, Kaiming He, Ross~B. Girshick, and Jian Sun.
\newblock Faster {R-CNN:} towards real-time object detection with region
  proposal networks.
\newblock {\em CoRR}, abs/1506.01497, 2015.

\bibitem{fcn}
Jonathan Long, Evan Shelhamer, and Trevor Darrell.
\newblock Fully convolutional networks for semantic segmentation.
\newblock {\em CoRR}, abs/1411.4038, 2014.

\bibitem{zhao_mrcnn}
Kang Zhao, Jungwon Kang, Jaewook Jung, and Gunho Sohn.
\newblock Building extraction from satellite images using mask r-cnn with
  building boundary regularization.
\newblock In {\em Proceedings of the IEEE Conference on Computer Vision and
  Pattern Recognition (CVPR) Workshops}, June 2018.

\bibitem{stiller_mrcnn}
Dorothee Stiller, Thomas Stark, Michael Wurm, Stefan Dech, and Hannes
  Taubenböck.
\newblock Large-scale building extraction in very high-resolution aerial
  imagery using mask r-cnn.
\newblock In {\em 2019 Joint Urban Remote Sensing Event (JURSE)}, pages 1--4,
  2019.

\bibitem{lstm}
Sepp Hochreiter and J\"{u}rgen Schmidhuber.
\newblock Long short-term memory.
\newblock {\em Neural Comput.}, 9(8):1735–1780, nov 1997.

\bibitem{unet_lstm_urbanchange}
Maria Papadomanolaki, Maria Vakalopoulou, and Konstantinos Karantzalos.
\newblock A deep multitask learning framework coupling semantic segmentation
  and fully convolutional lstm networks for urban change detection.
\newblock {\em IEEE Transactions on Geoscience and Remote Sensing},
  59(9):7651--7668, 2021.

\bibitem{conv_lstm_landuse}
Yue Zhu, Christian Geiß, Emily So, and Ying Jin.
\newblock Multitemporal relearning with convolutional lstm models for land use
  classification.
\newblock {\em IEEE Journal of Selected Topics in Applied Earth Observations
  and Remote Sensing}, 14:3251--3265, 2021.

\bibitem{geonames}
GeoNames Team.
\newblock Geonames.
\newblock \url{https://www.geonames.org/}, 2022.
\newblock Accessed: September 2021.

\bibitem{DG}
Maxar~Technologies Inc.
\newblock Digitalglobe.
\newblock \url{http://evwhs.digitalglobe.com/myDigitalGlobe}, 2022.
\newblock Accessed: September-December 2021.

\bibitem{geoboundaries}
Daniel Runfola, Austin Anderson, Heather Baier, Matt Crittenden, Elizabeth
  Dowker, Sydney Fuhrig, Seth Goodman, Grace Grimsley, Rachel Layko, Graham
  Melville, Maddy Mulder, Rachel Oberman, Joshua Panganiban, Andrew Peck, Leigh
  Seitz, Sylvia Shea, Hannah Slevin, Rebecca Youngerman, and Lauren Hobbs.
\newblock geoboundaries: A global database of political administrative
  boundaries.
\newblock {\em PLOS ONE}, 15(4):1--9, 04 2020.

\bibitem{luminance_convert}
MathWorks.
\newblock rgb2ntsc - convert rgb color values to ntsc color space.
\newblock \url{https://www.mathworks.com/help/images/ref/rgb2ntsc.html}, 2022.
\newblock Accessed: June 2022.

\bibitem{coco}
Tsung{-}Yi Lin, Michael Maire, Serge~J. Belongie, Lubomir~D. Bourdev, Ross~B.
  Girshick, James Hays, Pietro Perona, Deva Ramanan, Piotr Doll{\'{a}}r, and
  C.~Lawrence Zitnick.
\newblock Microsoft {COCO:} common objects in context.
\newblock {\em CoRR}, abs/1405.0312, 2014.

\bibitem{spacenet_shanghai}
SpaceNet.
\newblock Spacenet aoi 4 – shanghai.
\newblock \url{https://spacenet.ai/shanghai/}, 2022.
\newblock Accessed: May 2022.

\bibitem{Mstfakts}
Mustafa Aktas.
\newblock Building-detection-maskrcnn.
\newblock
  \url{https://github.com/Mstfakts/Building-Detection-MaskRCNN/blob/master/SpaceNet_train.py},
  2021.
\newblock Accessed: February 2022.

\bibitem{torchgeo}
Adam~J. Stewart, Caleb Robinson, Isaac~A. Corley, Anthony Ortiz, Juan
  M.~Lavista Ferres, and Arindam Banerjee.
\newblock Torchgeo: deep learning with geospatial data.
\newblock {\em CoRR}, abs/2111.08872, 2021.

\bibitem{ntl_urbanmapping}
Christopher Small, Francesca Pozzi, and C.D. Elvidge.
\newblock Spatial analysis of global urban extent from dmsp-ols night lights.
\newblock {\em Remote Sensing of Environment}, 96(3):277--291, 2005.

\bibitem{ni_ntl_poverty}
Ye~Ni, Xutao Li, Yunming Ye, Yan Li, Chunshan Li, and Dianhui Chu.
\newblock An investigation on deep learning approaches to combining nighttime
  and daytime satellite imagery for poverty prediction.
\newblock {\em IEEE Geoscience and Remote Sensing Letters}, 18(9):1545--1549,
  2021.

\bibitem{VIIRS}
Earth Observation Group~(EOG) at~Colorado School~of Mines.
\newblock Visible and infrared imaging suite (viirs).
\newblock \url{https://eogdata.mines.edu/products/vnl/}, 2022.
\newblock Accessed: June 2022.

\bibitem{shanghai_tearline}
Made in china 2025 and shanghai's zhangjiang high-tech industrial park.
\newblock
  \url{https://www.tearline.mil/public_page/tech-park-shanghai/#article_authors},
  2022.
\newblock Accessed: July 2022.

\bibitem{maskrcnn_high_AP}
Hao Su, Shunjun Wei, Min Yan, Chen Wang, Jun Shi, and Xiaoling Zhang.
\newblock Object detection and instance segmentation in remote sensing imagery
  based on precise mask r-cnn.
\newblock In {\em IGARSS 2019 - 2019 IEEE International Geoscience and Remote
  Sensing Symposium}, pages 1454--1457, 2019.

\end{thebibliography}

\end{document}